\documentclass[
a4paper,
man,
british
]{apa7}

\usepackage{changes}
\usepackage{cancel}
\usepackage[british]{babel}
\usepackage[utf8]{inputenc}
\usepackage{epstopdf}
\usepackage{csquotes}
\usepackage[flushleft]{threeparttable}
\usepackage{multirow}
\usepackage[
style = apa,
backend = biber,
sortcites = true,
sorting = nyt,
hyperref = true,
backref = false,
]{biblatex}
\hypersetup{
colorlinks = true,
linkcolor = black,
anchorcolor = black,
citecolor = black,
filecolor = black,
urlcolor = blue
}
\usepackage{float}
\usepackage{placeins}
\usepackage{xcolor}
\usepackage[toc, page]{appendix}
\usepackage{lscape}
\usepackage{afterpage}
\usepackage{esvect}
\usepackage{amsmath}
\usepackage{ragged2e}
\justifying\let\raggedright\justifying
\usepackage{enumitem}
\usepackage{makecell}
\DeclareLanguageMapping{british}{british-apa}
\usepackage[nodisplayskipstretch]{setspace}
\usepackage{subcaption} 
\usepackage{rotating}
\usepackage{geometry}
\usepackage{nameref}
\let \cite \parencite

\newcommand{\figurehere}[1]{\begin{center}%
\vspace{-2mm}
=========================\\%
Insert Figure #1 about here\\%
=========================\\%
\vspace{-2mm}
\end{center}}
\newcommand{\tablehere}[1]{\begin{center}%
\vspace{-2mm}
=========================\\%
Insert Table #1 about here\\%
=========================\\%
\vspace{-2mm}
\end{center}}

\usepackage{array}
\newcommand{\PreserveBackslash}[1]{\let\temp=\\#1\let\\=\temp}
\newcolumntype{C}[1]{>{\PreserveBackslash\centering}p{#1}}
\newcolumntype{R}[1]{>{\PreserveBackslash\raggedleft}p{#1}}
\newcolumntype{L}[1]{>{\PreserveBackslash\raggedright}p{#1}}

\title{Efficient or Powerful? Trade-offs Between Machine Learning and Deep Learning for Mental Illness Detection on Social Media}
\shorttitle{ML vs. DL: Trade-offs in Social Media Mental Illness Detection}

\authorsnames[2,2,1,5,1,5,3,4]{%
  Zhanyi Ding, Zhongyan Wang, Yeyubei Zhang, Yuchen Cao, Yunchong Liu, Xiaorui Shen, Yexin Tian, Jianglai Dai
}
\authorsaffiliations{%
  {University of Pennsylvania, School of Engineering and Applied Science}, 
  {New York University, Center for Data Science}, 
  {Georgia Institute of Technology, College of Computing}, 
  {University of California, Berkeley, Department of EECS}, 
  {Northeastern University, Khoury College of Computer Science}
}

\abstract{Social media platforms provide valuable insights into mental health trends by capturing user-generated discussions on conditions such as depression, anxiety, and suicidal ideation. Machine learning (ML) and deep learning (DL) models have been increasingly applied to classify mental health conditions from textual data, but selecting the most effective model involves trade-offs in accuracy, interpretability, and computational efficiency. This study evaluates multiple ML models, including logistic regression, random forest, and LightGBM, alongside deep learning architectures such as ALBERT and Gated Recurrent Units (GRUs), for both binary and multi-class classification of mental health conditions. Our findings indicate that ML and DL models achieve comparable classification performance on medium-sized datasets, with ML models offering greater interpretability through variable importance scores, while DL models are more robust to complex linguistic patterns. Additionally, ML models require explicit feature engineering, whereas DL models learn hierarchical representations directly from text. Logistic regression provides the advantage of capturing both positive and negative associations between features and mental health conditions, whereas tree-based models prioritize decision-making power through split-based feature selection. This study offers empirical insights into the advantages and limitations of different modeling approaches and provides recommendations for selecting appropriate methods based on dataset size, interpretability needs, and computational constraints.}


\authornote{
Correspondence concerning this article should be addressed to Yuchen Cao, Northeastern University, E-mail: cao.yuch@northeastern.edu}

\begin{document}
\maketitle

\section{Introduction}\label{sec:intro}
Social media has emerged as a vital platform for understanding mental health trends, offering researchers access to large-scale, real-time textual data reflecting personal experiences, emotional states, and psychological distress. Given the vast volume of user-generated content, researchers have increasingly turned to machine learning (ML) and deep learning (DL) approaches to automate mental health detection, leveraging natural language processing (NLP) techniques for feature extraction and classification. Platforms such as Twitter, Reddit, and Facebook have become key sources for analyzing mental health discussions, motivating the development of ML and DL models for early identification of psychological conditions.

Mental illnesses affect approximately one in eight individuals globally, with depression alone impacting over 280 million people \cite{who2023depression}. Early detection of these conditions is crucial for timely intervention, yet traditional diagnostic methods—such as clinical assessments and self-reported surveys—are resource-intensive and lack real-time insights \cite{kessler2017trauma}. Analyzing social media data presents an alternative, data-driven approach for mental health monitoring, enabling scalable detection of distress signals and behavioral patterns \cite{Guntuku2017detecting, dechoudhury2013social}. Advances in artificial intelligence (AI) and NLP have facilitated the application of ML and DL techniques for mental health classification, demonstrating promising results in various studies \cite{Shatte2019ML}.

Despite these advancements, several challenges remain. The effectiveness of ML and DL models is often hindered by dataset biases, inconsistencies in preprocessing techniques, and the reliance on imbalanced training data, all of which affect model generalizability \cite{cao2024mental, Hargittai2015, Helmy2024}. Linguistic complexities—such as informal language, sarcasm, and context-dependent meanings—further complicate the accurate detection of mental health conditions in social media text \cite{calvo2017nlp}. Another critical issue is the trade-off between model performance and interpretability. Traditional ML models, such as logistic regression and random forests, provide interpretability through feature importance scores but may struggle with nuanced language understanding. In contrast, DL models, including transformer-based architectures (e.g., Bidirectional Encoder Representations from Transformers, BERT) and recurrent neural networks (e.g., Gated Recurrent Units, GRUs), excel at capturing linguistic patterns but function as black-box models, limiting transparency in decision-making.

While prior systematic reviews have explored ML and DL applications in mental health detection \cite{cao2024mental, liu2024systematic, chen2025yearoveryeardevelopmentsfinancialfraud}, there remains a need for an empirical evaluation that systematically compares model performance and interpretability across different classification tasks. This study addresses this gap by assessing ML and DL models in both binary and multiclass mental health classification settings using a publicly available dataset from Kaggle. The dataset includes various mental health conditions, such as depression, anxiety, stress, suicidal ideation, bipolar disorder, and personality disorders. Model performance is evaluated using weighted F1 score and area under the receiver operating characteristic curve (AUROC) to account for class imbalance. Additionally, we assess model interpretability through feature importance measures, including logistic regression coefficients, random forest Gini impurity reduction, and LightGBM gain-based ranking.

By examining the trade-offs between model accuracy, interpretability, and computational efficiency, this study provides empirical insights into selecting appropriate models for mental health classification on social media. The remainder of this paper is organized as follows. Section \nameref{sec:methods} describes the methodological framework, including data preparation, model development, and evaluation metrics. Section \nameref{sec:results} presents findings on dataset characteristics, model performance evaluation, and interpretability assessments. Finally, the Discussion and Conclusion sections summarize key insights, implications for mental health research, and directions for future work.

\section{Method}\label{sec:methods}
This section outlines the methodological framework of our study, covering data collection, preprocessing, model construction, and evaluation metrics. All experiments were conducted using Python~3, leveraging key libraries such as \texttt{pandas} for data processing, \texttt{scikit-learn} and \texttt{lightgbm} for traditional machine learning, \texttt{PyTorch} for deep learning, and \texttt{Transformers} for utilizing pre-trained language models. These tools facilitated efficient data handling, systematic hyperparameter tuning, and rigorous performance evaluation. All models were trained on Google Colab, utilizing a high-RAM configuration powered by an NVIDIA T4 GPU, which provided the computational efficiency required for computational tasks, especially DL models. The following sections detail each stage of our approach. Complete code for data preparation, model development, and evaluation is available on GitHub (the link will be provided upon acceptance).

\subsection{Data Preparation}\label{method:data}
An extensive and varied dataset is fundamental for effective mental health detection via machine learning. We employed the `Sentiment Analysis for Mental Health' dataset available on \href{https://www.kaggle.com/datasets/suchintikasarkar/sentiment-analysis-for-mental-health/data}{Kaggle}. This dataset amalgamates textual data from multiple sources that cover topics such as depression, anxiety, stress, bipolar disorder, personality disorders, and suicidal ideation. Data were primarily obtained from social media platforms like Reddit, Twitter, and Facebook, where individuals discuss personal experiences and mental health challenges. The data acquisition process involved using platform-specific APIs and web scraping, followed by removing duplicates, filtering out spam or irrelevant content, and standardizing mental health labels. Personal identifiers were also removed to adhere to ethical standards, resulting in a well-structured CSV file with unique identifiers for each entry. Despite its diversity, the dataset's varying demographics and language styles (e.g., slang and colloquialisms) present challenges for natural language processing. Our preprocessing pipeline was specifically designed to address these variations and balance class distributions as needed.

We applied a consistent preprocessing pipeline to ready the dataset for both traditional and deep learning models. Initially, we cleaned the text by removing extraneous elements such as URLs, HTML tags, mentions, hashtags, special characters, and extra whitespace. The text was then converted to lowercase to maintain consistency. Next, we removed common stopwords using the NLTK stopword list \cite{bird2009natural} to eliminate non-informative words. Finally, lemmatization was used to reduce words to their base forms, ensuring that different forms of a word are treated uniformly. The processed dataset was randomly split into training, validation, and test sets, with 20\% allocated for testing. The remaining data was further divided into training (75\%) and validation (25\%) sets to ensure reproducibility and optimize model tuning. 

For classification, the dataset labels were structured in two distinct ways. In the multi-class scenario, the original labels in the Kaggle dataset were directly used, consisting of six categories: Normal, Depression, Suicidal, Anxiety, Stress, and Personality Disorder. For binary classification, all non-Normal categories were grouped under a single `Abnormal' label.

In natural language processing, feature extraction depends on the model type. Traditional ML models require structured numerical representations, while DL models can process raw text sequences or dense vector embeddings.

For ML models, text is commonly converted into numerical features using techniques such as the bag-of-words (BoW) model \cite{harris1954distributional}, which represents documents as token count vectors but treats all words equally. To address this limitation, Term Frequency-Inverse Document Frequency (TF-IDF) \cite{jones1972statistical} enhances BoW by weighting words based on their importance—emphasizing informative terms while downplaying common ones. In this study, we employed TF-IDF vectorization to extract numerical features, incorporating unigrams and bigrams and limiting the feature space to 1,000 features to optimize computational efficiency and mitigate overfitting.

\subsection{Model Development}
A variety of machine learning and deep learning models were developed to analyze and classify mental health statuses based on textual input. Each model was selected to capture different aspects of the data, ranging from simple linear classifiers to complex non-linear relationships. The following subsections outline the methodology of each model and its performance in both binary and multi-class classification.

\subsubsection{Logistic Regression}
Logistic regression is a fundamental classification technique widely used in social science and biomedical research \cite{hosmer2000applied}. It models the probability of a categorical outcome based on a weighted linear combination of input features. Despite its simplicity, logistic regression is still effective when applied to high-dimensional data, such as term frequency-based representations in natural language processing.

In this study, logistic regression served as an interpretable model that integrated various predictors (e.g., term frequencies) to estimate the probability of different mental health outcomes. The binary model predicts the likelihood of a positive case, while the multi-class extension accommodates multiple categories.

To prevent overfitting, model parameters were optimized using cross-entropy loss with regularization. A grid search was employed to fine-tune hyperparameters, including regularization strength, solver selection, and class weights, with the weighted F1 score guiding the selection process. The logistic regression models were implemented using the \texttt{LogisticRegression} class from \texttt{scikit-learn}.

\subsubsection{Support Vector Machine (SVM)}
Support Vector Machines (SVMs) are effective classifiers that identify an optimal decision boundary (hyperplane) to maximize the margin between classes \cite{cortes1995support}. Unlike probabilistic models such as logistic regression, SVMs utilize kernel functions to map input data into higher-dimensional spaces, allowing them to model both linear and non-linear relationships. Given the high-dimensional and sparse nature of text-based feature representations, both linear SVMs and non-linear SVMs with a radial basis function (RBF) kernel were evaluated, with model selection based on the weighted F1 score. Hyperparameter optimization was conducted via grid search, including regularization strength, class weighting, and $\gamma$ for RBF kernels\footnote{The gamma parameter determines the influence of individual training samples, where higher values result in more localized decision boundaries, while lower values promote broader generalization.}.

The final models were implemented using the \texttt{SVC} class from \texttt{scikit-learn}. For multi-class classification, the One-vs-One (OvO) strategy was employed, the default approach in \texttt{SVC}, which constructs pairwise binary classifiers for each class combination, with the final label determined through majority voting.

\subsubsection{Tree-Based Models}
Classification and Regression Trees (CART) are widely used for categorical outcome prediction in classification tasks. The algorithm constructs a binary decision tree by recursively partitioning the dataset based on predictor variables, selecting splits that optimize a predefined criterion. Common impurity measures, such as Gini impurity and entropy, assess split quality, with lower values indicating greater homogeneity within a node \cite{bishop2006pattern}. The tree expands iteratively until stopping conditions, such as a minimum node size, maximum depth, or impurity reduction threshold, are met.

To prevent overfitting, pruning techniques \cite{breiman1984classification} reduce tree complexity by removing splits with minimal predictive value, enhancing generalizability. However, standalone CART models often overfit, making them less suitable for complex classification tasks. Instead, this study employed ensemble methods, such as Random Forests and Gradient Boosted Trees, to improve robustness and predictive performance.

\paragraph{Random Forests}
Random Forests aggregate multiple decision trees to enhance classification performance. Each tree is trained on a bootstrap sample, ensuring diversity, while a random subset of features is considered at each split to reduce correlation and improve generalization \cite{breiman2001random, zhang2025tutorialusingmachinelearning}. Unlike individual trees, Random Forests do not require pruning, with complexity managed through hyperparameters such as the number of trees, tree depth, and minimum sample requirements.

Hyperparameter tuning via grid search optimized the number of estimators, tree depth, and minimum split criteria, using the weighted F1 score as the primary evaluation metric to address class imbalance. The best-performing binary classification model effectively distinguished between Normal and Abnormal mental health statuses. For multi-class classification, the same hyperparameter grid was used with a refined search scope for efficiency, ensuring balanced classification performance across mental health categories.

Beyond predictive accuracy, feature importance analysis provided insights into key variables influencing classification decisions, enhancing model interpretability. Random Forest models were implemented using \texttt{RandomForestClassifier} from \texttt{scikit-learn}, with hyperparameter tuning via grid search on the validation set. 

\paragraph{Light Gradient Boosting Machine (LightGBM)}
LightGBM is an optimized gradient-boosting framework designed for efficiency and scalability, particularly in high-dimensional datasets. Unlike traditional Gradient Boosting Machines (GBMs), which sequentially refine predictions by correcting errors from prior models, LightGBM employs a leaf-wise tree growth strategy, enabling deeper splits in dense regions for improved performance \cite{ke2017lightgbm}. Additionally, histogram-based feature binning reduces memory usage and accelerates training, making LightGBM faster and more resource-efficient than standard GBMs \cite{friedman2001greedy}.

Grid search was used to optimize hyperparameters, including the number of boosting iterations, learning rate, tree depth, number of leaves, and minimum child samples. To address class imbalance, the class weighting parameter was tested with both `balanced` and `None` options. Model selection was guided by the weighted F1 score, ensuring balanced classification performance.

For binary classification, LightGBM effectively distinguished between Normal and Abnormal statuses. For multi-class classification, it predicted categories including Normal, Depression, Anxiety, and Personality Disorder. Evaluation metrics included precision, recall, F1 scores, confusion matrices, and one-vs-rest ROC curves. LightGBM's built-in feature importance analysis further enhanced interpretability by identifying key predictors. The models were implemented using \texttt{LGBMClassifier} from the \texttt{lightgbm} library, with hyperparameter tuning via grid search on the validation set.

\subsubsection{A Lite Version of Bidirectional Encoder Representations from Transformers (ALBERT)}
ALBERT \cite{lan2020albert} is an optimized variant of BERT \cite{devlin2019bert} designed to enhance computational efficiency while preserving strong NLP performance. It achieves this by employing parameter sharing across layers and factorized embedding parameterization, significantly reducing the total number of model parameters. Additionally, ALBERT introduces Sentence Order Prediction (SOP) as an auxiliary pretraining task to improve sentence-level coherence. These architectural refinements make ALBERT a computationally efficient alternative to BERT, particularly well-suited for large-scale text classification applications such as mental health assessment.

In this study, ALBERT was fine-tuned for both binary and multi-class classification. The binary model was trained to differentiate between Normal and Abnormal mental health statuses, while the multi-class model classified inputs into categories such as Normal, Depression, Anxiety, and Personality Disorder. The pretrained \texttt{Albert-base-v2} model was utilized, and hyperparameter optimization was conducted using random search over 10 iterations, tuning learning rates, dropout rates, and training epochs. Model performance was evaluated using the weighted F1 score as the primary metric. For the multi-class task, the classification objective was adjusted to predict seven categories, with weighted cross-entropy loss applied to address class imbalances.

ALBERT's architecture effectively captures long-range dependencies in text while offering substantial computational advantages. Performance optimization was conducted using random hyperparameter tuning within the Hugging Face \texttt{Transformers} framework, leveraging \texttt{AlbertTokenizer} and \texttt{AlbertForSequenceClassification} for implementation.

\subsubsection{Gated Recurrent Units (GRUs)}
Gated Recurrent Units (GRUs) are a variant of recurrent neural networks (RNNs) designed to model sequential dependencies, making them well-suited for natural language processing tasks such as text classification \cite{cho2014learning}. Compared to Long Short-Term Memory networks (LSTMs), GRUs provide greater computational efficiency by simplifying the gating mechanism. Specifically, they merge the forget and input gates into a single update gate, reducing the number of parameters while effectively capturing long-range dependencies.

In this study, GRUs were employed for both binary and multi-class mental health classification. The binary model differentiated between Normal and Abnormal mental health statuses, while the multi-class model predicted categories such as Normal, Depression, Anxiety, and Personality Disorder.

The GRU architecture consisted of three primary components:
\begin{itemize}
  \item \textbf{Embedding Layer}: Maps token indices to dense vector representations of a fixed size.
  \item \textbf{GRU Layer}: Processes sequential inputs, preserving contextual dependencies, with the final hidden state serving as the input to the classifier.
  \item \textbf{Fully Connected Layer}: Transforms the hidden state into output logits corresponding to the classification categories.
\end{itemize}
To mitigate overfitting, dropout regularization was applied, and weighted cross-entropy loss was used to address class imbalance.

Hyperparameter tuning was conducted via random search, optimizing key parameters such as embedding dimensions, hidden dimensions, learning rates, and training epochs. The weighted F1 score was used for model selection, ensuring robust performance on both validation and test data.

Overall, GRUs effectively captured sequential patterns in text, enabling the extraction of linguistic features relevant to mental health classification. While less interpretable than tree-based models, their efficiency and ability to model long-range dependencies make them well-suited for text classification. The models were implemented using PyTorch's \texttt{torch.nn} module, incorporating \texttt{nn.Embedding}, \texttt{nn.GRU}, and \texttt{nn.Linear} layers. Optimization was performed using \texttt{torch.optim.Adam}, with class imbalances handled through \texttt{nn.CrossEntropyLoss}.

\subsection{Evaluation Metrics}
Classifying mental health conditions, such as depression or suicidal ideation, often involves imbalanced class distributions, where the `positive' class (e.g., individuals experiencing a mental health condition) is significantly underrepresented compared to the `negative' class (e.g., no reported issues). In such cases, traditional metrics like accuracy can be misleading, as a model predicting only the majority class may still achieve high accuracy despite failing to detect minority-class cases. To provide a more comprehensive assessment of classification performance, the following evaluation metrics were used:

\begin{itemize}
  \item \textbf{Recall (Sensitivity)}: Captures the proportion of actual positive cases correctly identified. High recall is crucial in mental health detection to minimize false negatives and ensure individuals in need receive appropriate intervention \cite{Bradford2024Diagnostic}. However, excessive focus on recall may increase false positives, leading to potential misclassifications.
  \item \textbf{Precision}: Measures the proportion of predicted positive cases that are actually positive. High precision is critical in mental health classification, as false positives can lead to unnecessary concern, stigma, and unwarranted interventions \cite{Bradford2024Diagnostic, Wei2023CancerGeneClassification}. However, optimizing for precision alone may cause the model to miss true positive cases, limiting its usefulness.
  \item \textbf{F1 Score}: Represents the harmonic mean of precision and recall, offering a balanced performance measure \cite{powers2011evaluation}. This metric is particularly useful for imbalanced datasets, ensuring that neither precision nor recall is disproportionately optimized at the expense of the other.
  \item \textbf{Area Under the Receiver Operating Characteristic Curve (AUROC)}: Assesses the model's ability to distinguish between positive and negative cases across various classification thresholds. Although AUROC provides an overall measure of discrimination performance, it may be less informative in severely imbalanced datasets, where the majority class dominates \cite{davis2006relationship, tao2024nevlpnoiserobustframeworkefficient}.
\end{itemize}

\section{Results}\label{sec:results}
This section presents the findings from the analysis of the dataset and the evaluation of machine learning and deep learning models for mental health classification. First, we provide an \textit{Overview of Mental Health Distribution}, highlighting the inherent class imbalances within the dataset and their implications for model development. Next, the \textit{Hyperparameter Optimization} subsection details the parameter tuning process, which ensures that each model performs at its best configuration for both binary and multi-class classification tasks. Finally, the \textit{Model Performance Evaluation} subsection compares the models' performance based on key metrics, including F1 scores and Area Under the Receiver Operating Characteristic Curve (AUROC). Additionally, nuanced observations, such as the challenges associated with underrepresented classes, are discussed to provide deeper insights into the modeling outcomes.

\subsection{Distribution of Mental Health Status}
The dataset contains a total of 52,681 unique textual statements, each annotated with a corresponding mental health status label. The labels represent various mental health categories, reflecting the distribution of conditions within the dataset. 

The dataset is heavily imbalanced, with certain categories having significantly higher representation than others. Specifically:
\vspace{-1.5mm}
\begin{itemize}
  \item \textbf{Normal}: 16,343 statements (31.02\%)
  \vspace{-3mm}
  \item \textbf{Depression}: 15,404 statements (29.24\%)
  \vspace{-3mm}
  \item \textbf{Suicidal}: 10,652 statements (20.22\%)
  \vspace{-3mm}
  \item \textbf{Anxiety}: 3,841 statements (7.29\%)
  \vspace{-3mm}
  \item \textbf{Bipolar}: 2,777 statements (5.27\%)
  \vspace{-3mm}
  \item \textbf{Stress}: 2,587 statements (4.91\%)
  \vspace{-3mm}
  \item \textbf{Personality Disorder}: 1,077 statements (2.04\%)
\end{itemize}
\vspace{-1.5mm}

For the binary classification task, all mental health conditions (Depression, Suicidal, Anxiety, Bipolar, Stress, and Personality Disorder) were combined into a single category labeled as \textbf{Abnormal}, while the \textbf{Normal} category remained unchanged. This transformation resulted in:
\vspace{-1.5mm}
\begin{itemize}
  \item \textbf{Normal}: 16,343 statements (31.02\%)
  \vspace{-3mm}
  \item \textbf{Abnormal}: 36,338 statements (68.98\%)
\end{itemize}
\vspace{-1.5mm}

Such imbalance feature in both multi-class and binary classification tasks highlights the importance of evaluation metrics that account for disparities, such as the weighted F1 score. 

\subsection{Computational Efficiency}
The computational time for training the models varied significantly based on the algorithm type and classification task. Among ML models, SVM required an exceptionally long training time, far exceeding other ML approaches like Logistic Regression, Random Forest, and Light GBM, for both binary and multi-class tasks. In contrast, DL models such as ALBERT and GRU consistently required more time compared to ML models, reflecting their higher computational complexity.

For ML models, training times for multi-class classification were longer than for binary classification, likely due to the increased complexity of predicting multiple categories. However, for DL models, there was no notable difference in training times between binary and multi-class tasks, indicating that their computational cost was primarily driven by model architecture rather than the number of classes.

A detailed information of training times is presented in Table~\ref{tab1:training_times}.

\tablehere{1}

\subsection{Performance Metrics}
Table~\ref{tab2:binary_metrics} presents the weighted F1 scores and AUROC values for all models evaluated on binary classification tasks. Across all models, there were minimal numerical differences in performance, with all achieving strong results in both metrics. The F1 scores ranged from 0.9345 (Logistic Regression) to 0.9576 (ALBERT), while AUROC values were consistently high, spanning from 0.92 (Random Forest) to 0.95 (ALBERT). These results indicate that all models effectively distinguished between Normal and Abnormal mental health statuses.

Despite the close performance across models, a general trend emerged where deep learning (DL) models, such as ALBERT and GRU, outperformed traditional machine learning (ML) models. For instance, ALBERT achieved the highest F1 score (0.9576) and AUROC (0.95), while GRU closely followed with an F1 score of 0.9512 and an AUROC of 0.94. In contrast, ML models such as Logistic Regression, Random Forest, and LightGBM showed slightly lower, albeit still competitive, performance.

Table~\ref{tab3:multiclass_metrics} summarizes the weighted F1 scores and micro-average AUROC values for multi-class classification tasks. Similar to binary classification, the differences in performance across models were small, with DL models generally outperforming ML models. ALBERT achieved the highest F1 score (0.7841) and shared the top AUROC value (0.97) with LightGBM and GRU. ML models such as Logistic Regression and Random Forest exhibited slightly lower F1 scores, at 0.7498 and 0.7478, respectively, but still demonstrated strong AUROC values (0.96).

\tablehere{2}

Notably, a consistent pattern was observed where multi-class classification yielded lower F1 scores compared to binary classification across all models. The lower F1 scores for multi-class classification compared to binary classification reflect the increased complexity of predicting seven distinct mental health categories. Binary classification requires only a single decision boundary between Normal and all other classes (combined into Abnormal), whereas multi-class classification must learn multiple boundaries between overlapping categories like Depression, Anxiety, and Stress. This added complexity introduces more opportunities for misclassification, further lowering F1 scores. On the contrary, the AUROC values remained consistently high for both binary and multi-class tasks, indicating robust discrimination between classes despite the added complexity.

\tablehere{3}

\begin{figure}[h!]
 \centering
 \includegraphics[width=0.8\textwidth]{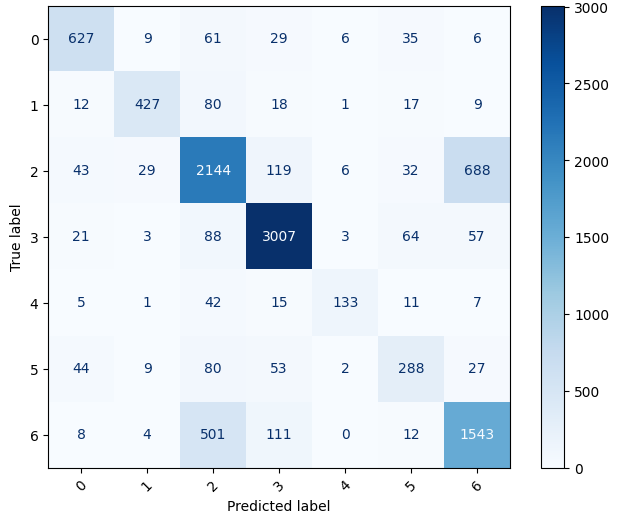}
 \caption{LightGBM Multi-Class Model Performance: Confusion Matrix. \\ The class labels are arranged as follows: 
 Class 0: Anxiety, Class 1: Normal, Class 2: Depression, 
 Class 3: Suicidal, Class 4: Stress, Class 5: Bipolar, and 
 Class 6: Personality Disorder.}
 \label{fig1:CM}
\end{figure}

The discrepancy between the F1 score and AUROC observed in the multi-class classification results can be attributed to the fundamental differences in what these metrics measure. The F1 score, which balances precision and recall, is sensitive to class imbalance and specific misclassifications. In the confusion matrix (Figure~\ref{fig1:CM}), generated for the LightGBM multi-class model and included here for illustration purposes, certain classes such as Suicidal (Class 6) and Depression (Class 2) show notable misclassifications, including frequent overlaps with Stress (Class 5) and Normal (Class 3). This directly impacts the F1 score by lowering the precision and recall for these specific classes.

\begin{figure}[h!]
 \centering
 \includegraphics[width=0.8\textwidth]{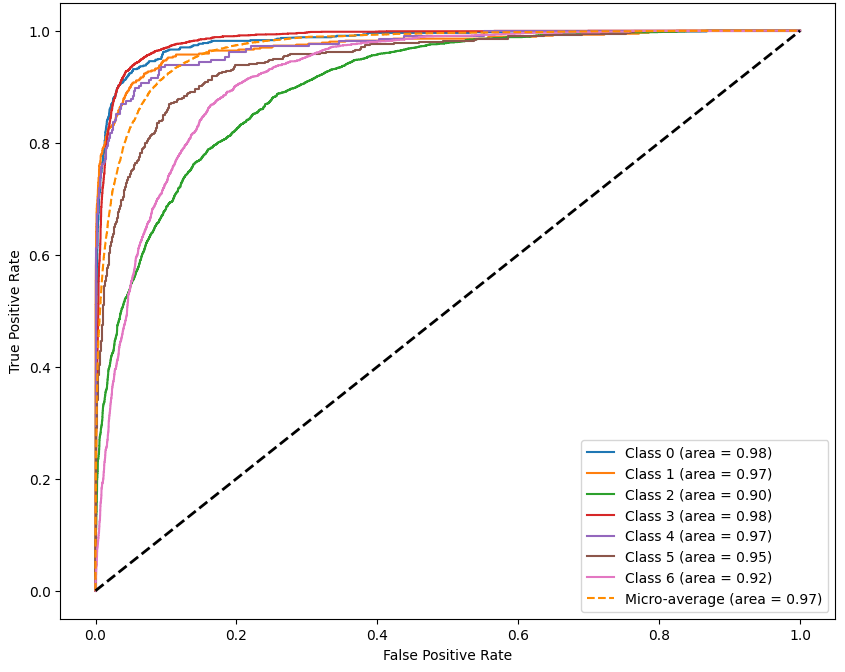}
 \caption{LightGBM Multi-Class Model Performance: Area Under the Receiver Operating Characteristic Curve. \\
 The class labels are arranged as follows: 
 Class 0: Anxiety, Class 1: Normal, Class 2: Depression, 
 Class 3: Suicidal, Class 4: Stress, Class 5: Bipolar, and 
 Class 6: Personality Disorder.}
 \label{fig2:AUROC}
\end{figure}

In contrast, AUROC measures the model's ability to rank predictions correctly across thresholds, and it remains robust to class imbalances and individual misclassification errors. The ROC curves (Figure~\ref{fig2:AUROC}), also from the LightGBM multi-class model and included for illustrative purposes, demonstrate strong separability for most classes, with areas under the curve (AUC) exceeding 0.90 for all but Class 2 (Depression) and Class 6 (Suicidal). The micro-average AUROC of 0.97 indicates that the model can effectively rank instances across all classes, even when specific misclassifications reduce the F1 score.

\subsection{Error Analyses}
The confusion matrix reveals specific patterns of misclassification that contribute to the lower F1 scores for some classes in the multi-class classification task. Key observations include:
\vspace{-1.5mm}
\begin{itemize}
    \item \textbf{Overlap Between Emotionally Similar Classes}: \textit{As indicated in Figure\ref{fig1:CM}, Depression (Class 2)} and \textit{Personality Disorder (Class 6)} show significant overlap, with many instances of Depression misclassified as Personality Disorder or vice versa. Similarly, \textit{Suicidal (Class 3)} was frequently misclassified as Depression, likely due to overlapping linguistic patterns. Another possible explanation lies in the nature of the dataset itself, which was constructed by combining data from multiple sources. While these labels may have been well-defined and effective for their original studies, they may lack consistency when integrated into a unified dataset, leading to ambiguity in class boundaries.
    \vspace{-3mm}
    \item \textbf{Poor Discrimination for Depression}: The ROC curve (Figure~\ref{fig2:AUROC}) highlights that \textit{Depression (Class 2)} has the lowest AUC (0.90) among all classes in the LightGBM model. For other models, the AUC for Class 2 drops even further, indicating consistent difficulty in distinguishing Depression from other classes. This is likely due to semantic overlap with related classes such as \textit{Stress (Class 4)}, \textit{Suicidal (Class 3)}, and \textit{Personality Disorder (Class 6)}. Additionally, inconsistencies in labeling across data sources may further exacerbate the challenge of identifying Depression accurately.
    \vspace{-3mm}
    \item \textbf{Underrepresented Classes and Data Imbalance}: \textit{Bipolar (Class 5)} and \textit{Personality Disorder (Class 6)} were underrepresented in the dataset, which exacerbated misclassification issues.
\end{itemize}
\vspace{-1.5mm}

\subsection{Model Interpretability}
In traditional machine learning (ML) models, variable importance can be quantified to understand how individual features contribute to predictions. This interpretability allows researchers to identify key linguistic and behavioral markers associated with mental health conditions. However, deep learning (DL) models operate differently. Rather than relying on explicit features, DL models extract representations from raw text, making them inherently black-box models. Since these models learn hierarchical patterns across entire sentences and contexts, they do not produce traditional variable importance scores, making direct interpretability more challenging. In this project, we assessed variable importance for three out of four machine learning models: logistic regression, random forest, and LightGBM. Support Vector Machine (SVM) was excluded from this analysis because Radial Basis Function (RBF) kernel was selected during model construction, which is a nonlinear kernel. In such cases, variable importance is not directly interpretable due to the transformation of the input space, making it difficult to quantify individual feature contributions meaningfully \cite{guyon2002gene}. Unlike linear models, where coefficients provide a direct measure of feature importance, nonlinear SVMs construct decision boundaries in high-dimensional spaces, where the contribution of each feature depends on complex interactions \cite{chang2011libsvm}.

For logistic regression, variable importance is derived from model coefficients, where positive coefficients indicate a higher likelihood of the outcome (e.g., mental health condition), while negative coefficients suggest a protective effect. To enhance interpretability, we adopted a color scheme in our visualizations: dark gray for positive coefficients and light gray for negative coefficients. For Random Forest, variable importance is computed using the Gini impurity reduction criterion \cite{breiman2001random}. This metric quantifies how much each feature contributes to reducing class impurity across the decision trees by assessing the decrease in Gini impurity at each node split. Features with higher importance scores have a greater impact on classification performance. For LightGBM, variable importance is measured using information gain, which quantifies the total improvement in the model's objective function when a feature is used for node splitting across all trees in the boosting process. Information gain reflects how much a feature contributes to minimizing the loss function during training and is commonly used in gradient boosting frameworks \cite{ke2017lightgbm}. Features with higher gain values contribute more to optimizing the model's predictive accuracy. 

\begin{figure}[htbp]
    \centering
    % Row 1
    \begin{subfigure}[b]{0.48\textwidth}
        \centering
        \includegraphics[width=\textwidth]{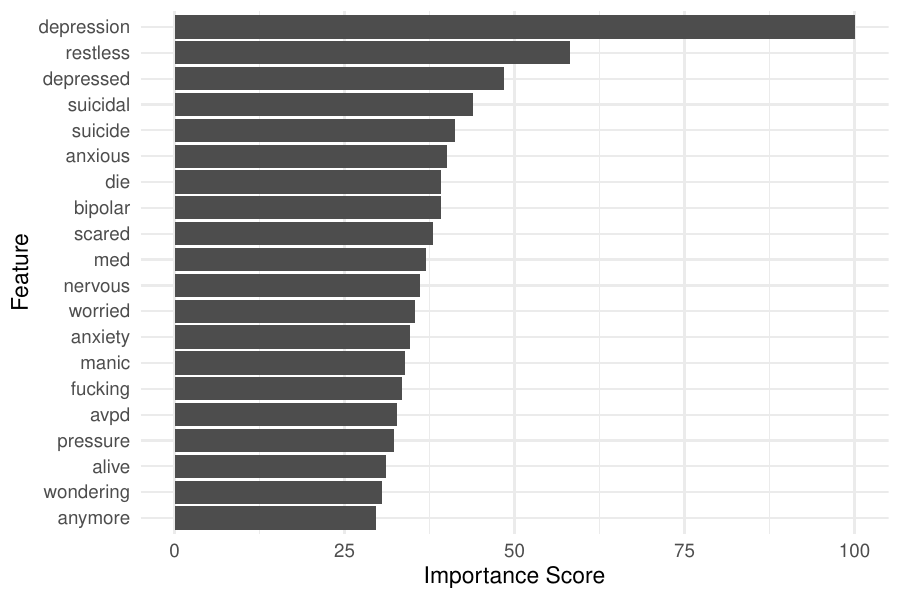}
        \caption{Logistic Regression (Binary)}
    \end{subfigure}
    \hfill
    \begin{subfigure}[b]{0.48\textwidth}
        \centering
        \includegraphics[width=\textwidth]{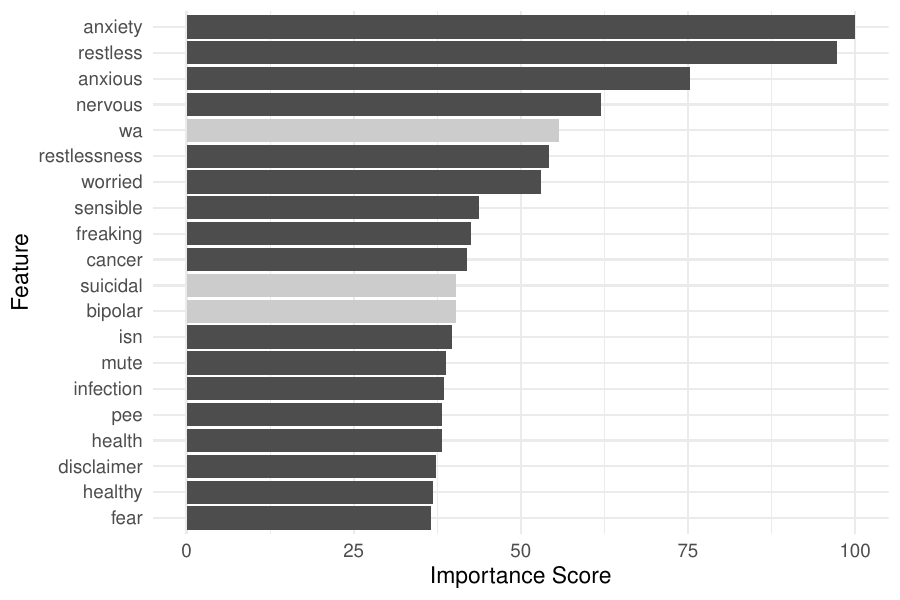}
        \caption{Logistic Regression (Multiclass)}
    \end{subfigure}

    % Row 2
    \begin{subfigure}[b]{0.48\textwidth}
        \centering
        \includegraphics[width=\textwidth]{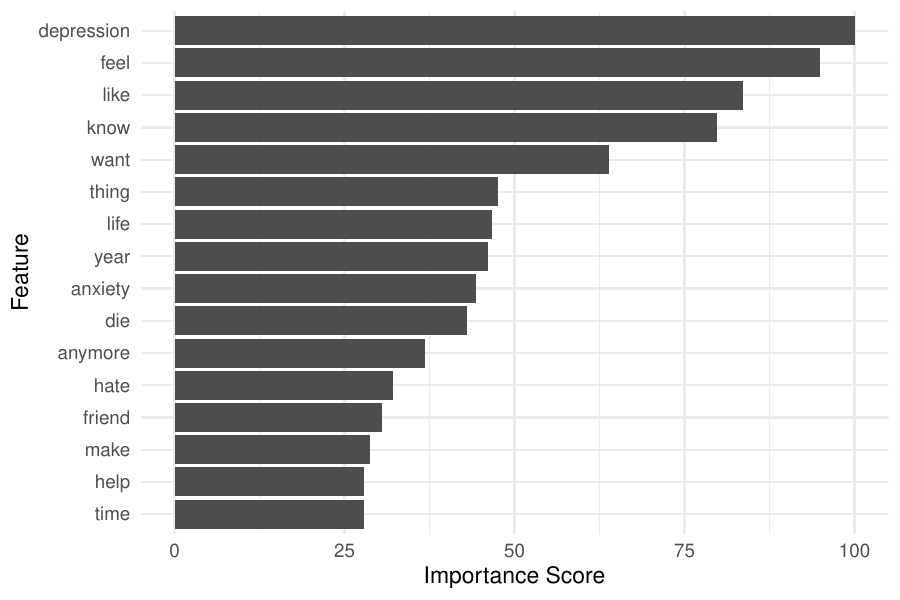}
        \caption{Random Forest (Binary)}
    \end{subfigure}
    \hfill
    \begin{subfigure}[b]{0.48\textwidth}
        \centering
        \includegraphics[width=\textwidth]{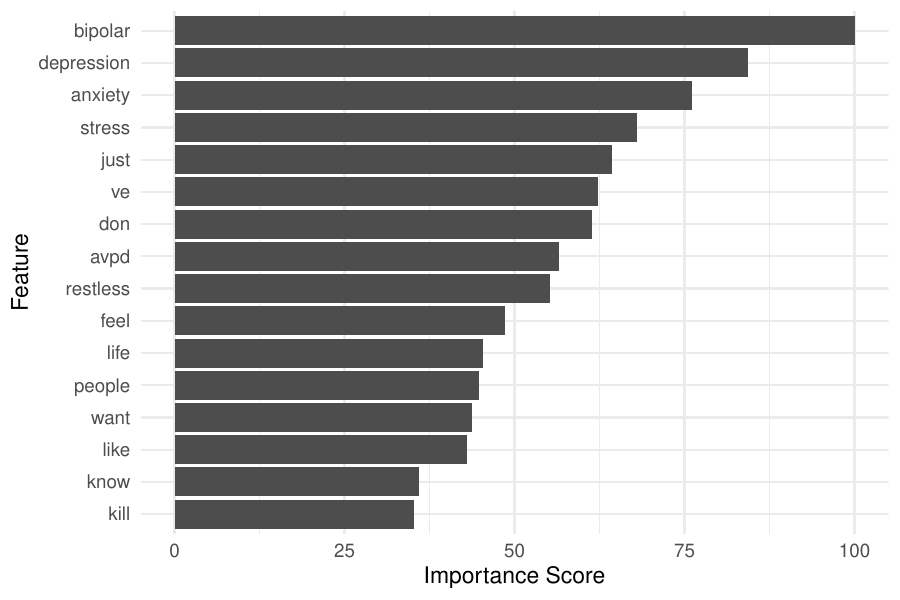}
        \caption{Random Forest (Multiclass)}
    \end{subfigure}
   
     % Row 3    
     \begin{subfigure}[b]{0.48\textwidth}
        \centering
        \includegraphics[width=\textwidth]{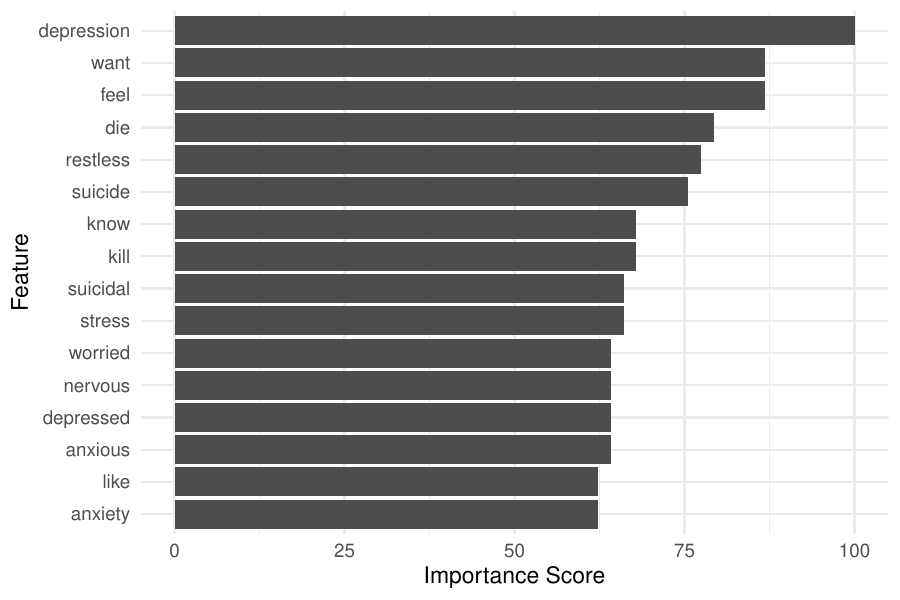}
        \caption{LightGBM (Binary)}
    \end{subfigure}
    \hfill
    \begin{subfigure}[b]{0.48\textwidth}
        \centering
        \includegraphics[width=\textwidth]{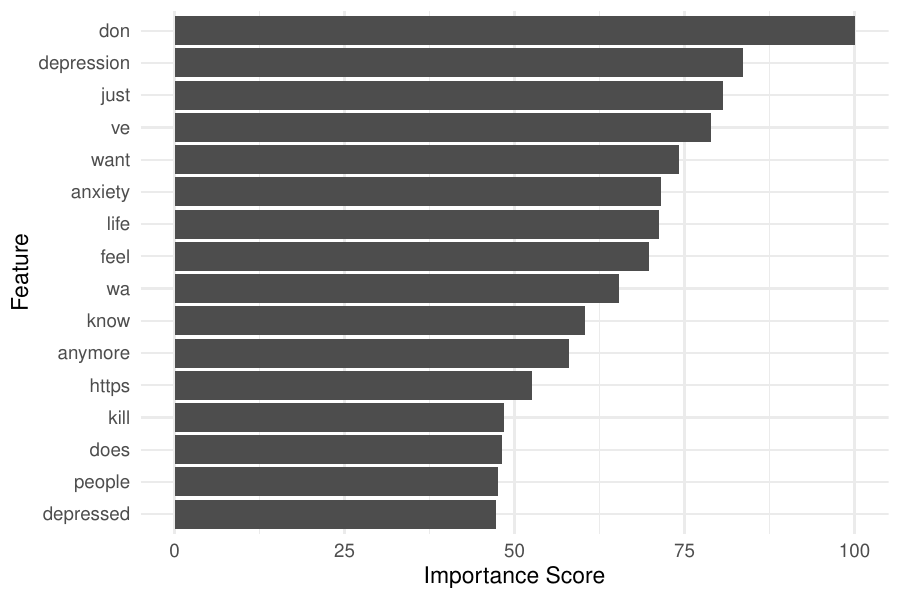}
        \caption{LightGBM (Multiclass)}
    \end{subfigure}
    \caption{Comparison of Feature Importance Across Different Models}
    \label{fig:feature_importance}
\end{figure}

The variable importance for both binary and multiclass models using logistic regression, random forest, and LightGBM is presented in Figure \ref{fig:feature_importance}. To ensure comparability across models, we rescaled the variable importance scores for random forest and LightGBM by normalizing them to a maximum value of 100. For logistic regression, variable importance is represented by model coefficients, retaining both their relative scale and sign. Among machine learning models that provide feature importance, logistic regression offers a more interpretable framework since its importance scores are derived directly from model coefficients. Unlike tree-based methods, which rely on splitting criteria (such as Gini impurity for Random Forest or Gain for LightGBM), logistic regression coefficients retain their sign, allowing researchers to distinguish positive and negative associations with the target outcome. This property is particularly valuable in mental health detection, where it is critical to understand whether a term increases or decreases the likelihood of classification (e.g., identifying depressive symptoms).

\figurehere{3}

Despite variations in ranking, the top features identified across machine learning models share strong overlap, reinforcing their relevance in mental health classification. However, different importance criteria lead to model-specific variations: logistic regression ranks features based on coefficient magnitude (allowing both positive and negative values), random forest uses Gini impurity reduction, and LightGBM employs gain-based ranking. While these models prioritize features differently, they consistently highlight depression-related language as the strongest predictor of mental health conditions on social media.

For binary classification models, `depression' emerges as the most predictive feature across all methods, reinforcing its centrality in identifying mental health status. Beyond this, words associated with emotional distress—such as `feel,' `want,' and `anxiety'—consistently appear in the top ranks, though their order varies. Logistic regression assigns strong positive coefficients to `restless' and `suicidal,' suggesting their direct correlation with depressive states. Meanwhile, tree-based models (random forest and LightGBM) highlight terms like `die,' `kill,' and `suicide' more prominently, likely due to their effectiveness in decision splits. These differences reflect how each model processes textual features, with logistic regression providing interpretability through sign-based coefficients, while tree-based models prioritize decision-making power through split-based feature selection.

In the multiclass setting, feature importance rankings shift to reflect the distinctions between different mental health conditions. While `depression' remains a dominant predictor, terms like `bipolar' and `anxiety' gain prominence, particularly in tree-based models (random forest and LightGBM), suggesting their utility in distinguishing among multiple mental health states. Logistic regression, on the other hand, highlights `restless' and `nervous' more strongly, aligning with its emphasis on anxiety-related symptoms. The presence of `kill' and `suicidal' in tree-based models underscores their role in severe mental health classifications. Despite these ranking differences, the core predictive features remain largely consistent, validating their role in mental health detection on social media.

Among models capable of generating variable importance, logistic regression stands out for its interpretability. Unlike tree-based methods, which assign importance based on split-based metrics, logistic regression allows for direct interpretation of feature coefficients, capturing both positive and negative associations. This provides a clearer understanding of which terms contribute most strongly to classification and in what direction. In contrast, while random forest and LightGBM effectively rank important features, their criteria for feature selection make direct interpretability more challenging.

\section{Discussion}
This study provides an empirical evaluation of machine learning (ML) and deep learning (DL) models for mental health classification on social media, focusing on their predictability, interpretability, and computational efficiency. The findings highlight key trade-offs that researchers should consider when selecting models for mental health detection tasks. While DL models, such as ALBERT and GRU, have gained popularity for their ability to extract hierarchical representations from raw text, their advantages in small-to-medium datasets remain limited. The results indicate that in cases where dataset size is moderate, traditional ML models, such as logistic regression, random forests, and LightGBM, perform comparably to DL models while offering additional benefits in terms of interpretability and computational efficiency.  

The size of the dataset plays a crucial role in determining the most suitable modeling approach. When working with small to medium-sized datasets, traditional ML models remain an effective choice. Their reliance on structured feature engineering, while requiring additional preprocessing efforts, allows for a more controlled and interpretable learning process. In contrast, DL models require large-scale training data to leverage their full potential. Although DL architectures can automatically extract complex linguistic patterns without extensive feature engineering, this advantage is less pronounced in settings with limited training samples. For researchers with small datasets, the use of feature engineering and careful selection of input variables is critical to optimizing model performance. The results suggest that DL models are more suitable for large-scale mental health detection tasks, where the volume of data is sufficient to justify their increased computational demands.  

In addition to dataset size, computational efficiency remains a practical consideration in model selection. The ML models evaluated in this study consistently required less computational time compared to DL models, making them preferable in scenarios where efficiency is a priority. While DL models demonstrated competitive classification performance, their significantly longer training times present a challenge, particularly for researchers working with constrained computing resources. Given that many mental health detection applications require scalable solutions, this finding suggests that ML models provide a more efficient and accessible alternative for researchers seeking to deploy classification models without extensive computational infrastructure.  

Interpretability is another critical factor in model selection, particularly for applications in mental health research where understanding the relationships between linguistic patterns and psychological states is essential. Among the ML models, logistic regression offers the clearest interpretability, as it provides direct coefficient estimates that indicate the relative influence of each feature. This advantage is particularly important in mental health classification, where identifying protective and risk-related linguistic markers can provide valuable insights into early detection and intervention strategies. Unlike logistic regression, tree-based models such as random forests and LightGBM do not distinguish between positive and negative associations but instead rank features based on their contribution to classification accuracy. This limitation reduces their interpretability but still allows for the identification of key predictive features. In contrast, DL models operate as black-box systems with no explicit feature importance scores, making them less suitable for researchers prioritizing explainability. Given these differences, logistic regression emerges as the preferred choice when interpretability is a key concern, while tree-based models provide flexibility for high-dimensional data without imposing strong linearity assumptions.  

Despite the strengths of ML models in terms of efficiency and interpretability, it is important to acknowledge the assumptions underlying these approaches. Logistic regression, for example, assumes a linear relationship between the features and the log-odds of the target variable, an assumption that was not explicitly tested in this study. Future research should explore whether nonlinear transformations or interaction terms could improve model fit while maintaining interpretability. Similarly, while tree-based models do not impose strict assumptions about feature relationships, they rely on hierarchical partitioning mechanisms that may introduce biases in highly unbalanced datasets. These limitations highlight the importance of methodological rigor when selecting ML models for mental health research.  

In addition to model selection, dataset composition and label consistency present challenges in mental health classification. The dataset used in this study was compiled from multiple publicly available sources, which, while beneficial for enhancing linguistic diversity, also introduced inconsistencies in class labels. Since each dataset was originally created for different research purposes, class boundaries may not be clearly defined when combined. This issue likely contributed to increased misclassification rates in the multi-class setting, particularly in categories with overlapping linguistic features such as depression, stress, and suicidal ideation. The presence of ambiguous class definitions suggests that future studies should consider collecting data directly from social media platforms using standardized labeling criteria. By ensuring greater consistency in data annotation, researchers can improve model generalizability and reduce classification errors.  

Another limitation relates to annotation quality. Given the subjective nature of mental health expressions, the reliability of pre-existing labels in publicly available datasets can be uncertain. Manual verification of labels by domain experts could improve classification accuracy, but such an approach is time-consuming and resource-intensive. As an alternative, future work could explore Artificial Intelligence-assisted annotation strategies to enhance labeling consistency. Advances in natural language processing, particularly in large language models, offer opportunities for developing semi-automated annotation systems that incorporate human-in-the-loop validation. By combining automated text classification with expert oversight, researchers could create more comprehensive and reliable datasets for mental health detection.  

The ethical implications of using social media data for mental health research also warrant careful consideration. While these datasets provide valuable insights into psychological well-being, they often include sensitive information that must be handled with caution. Privacy-preserving techniques, such as anonymization and differential privacy, should be explored to protect user identities while maintaining the linguistic features necessary for classification. Future research should also establish clearer guidelines for ethical data collection, ensuring that social media-derived datasets align with best practices in mental health ethics.  

In summary, this study provides a comparative analysis of ML and DL models for mental health classification on social media, highlighting key considerations in accuracy, interpretability, and computational efficiency. The findings suggest that ML models remain a practical and interpretable choice for small to medium-sized datasets, while DL models may offer advantages when working with larger data volumes. Among ML models, logistic regression is particularly useful for its ability to distinguish between positive and negative feature importance, offering valuable insights into linguistic markers associated with mental health conditions. However, researchers should remain mindful of model assumptions and dataset inconsistencies, which can impact classification performance. Moving forward, efforts to improve data collection, annotation quality, and ethical considerations will be essential for advancing AI-driven mental health detection and ensuring that these models contribute to more effective, transparent, and responsible research practices.

\printbibliography

\newpage
\begin{table}[h!]
\centering
\begin{tabular}{l|l|c|c}
\hline
\textbf{Type}  & \textbf{Model}  & \textbf{Binary (seconds)} & \textbf{Multiclass (seconds)} \\ \hline
\multirow{4}{*}{ML}  & SVM & 4681.96  & 22844.23 \\ \cline{2-4} 
 & Logistic Regression & 7.33 & 181.86  \\ \cline{2-4} 
 & Random Forest & 263.54 & 2895.43 \\ \cline{2-4} 
 & Light GBM & 336.65 & 3968.33  \\ \hline
\multirow{2}{*}{DL}  & Albert  & 21244.18 & 20860.15  \\ \cline{2-4} 
 & GRU & 1530.76  & 1567.24 \\ \hline
\end{tabular}
\caption{Training times (in seconds) for model optimization in binary and multi-class classification tasks}
\label{tab1:training_times}
\end{table}

\begin{table}[h!]
\centering
\begin{tabular}{l|c|c}
\hline
\textbf{Model}  & \textbf{F1 Score} & \textbf{AUROC} \\ \hline
Support Vector Machine (SVM) & 0.9401  & 0.93 \\ \hline
Logistic Regression  & 0.9345  & 0.93 \\ \hline
Random Forest  & 0.9359  & 0.92 \\ \hline
LightGBM & 0.9358  & 0.93 \\ \hline
ALBERT & 0.9576  & 0.95 \\ \hline
Gated Recurrent Units (GRU)  & 0.9512  & 0.94 \\ \hline
\end{tabular}
\caption{F1 Scores and AUROC for Binary Classification Tasks.}
\label{tab2:binary_metrics}
\end{table}

\begin{table}[h!]
\centering
\begin{tabular}{l|c|c}
\hline
\textbf{Model} & \textbf{F1 Score} & \textbf{Micro-Average AUROC} \\ \hline
Support Vector Machine (SVM) & 0.7610  & 0.95 \\ \hline
Logistic Regression  & 0.7498  & 0.96 \\ \hline
Random Forest  & 0.7478  & 0.96 \\ \hline
LightGBM & 0.7747  & 0.97 \\ \hline
ALBERT & 0.7841  & 0.97 \\ \hline
Gated Recurrent Units (GRU)  & 0.7756  & 0.97 \\ \hline
\end{tabular}
\caption{F1 Scores and Micro-Average AUROC for Multi-Class Classification Tasks.}
\label{tab3:multiclass_metrics}
\end{table}

\newpage

\end{document}